\documentclass[10pt,twocolumn,letterpaper]{article}

\usepackage{cvpr}              

\usepackage[dvipsnames]{xcolor}
\usepackage{bm}

\definecolor{cvprblue}{rgb}{0.21,0.49,0.74}
\usepackage[pagebackref,breaklinks,colorlinks,citecolor=cvprblue,linkcolor=cvprblue]{hyperref}

\title{Improving Subject-Driven Image Synthesis with Subject-Agnostic Guidance}

\author{Kelvin C.K. Chan \quad Yang Zhao \quad Xuhui Jia \quad Ming-Hsuan Yang \quad Huisheng Wang\\\\
Google
}

\begin{document}

\twocolumn[{%
	\renewcommand\twocolumn[1][]{#1}%
	\maketitle\thispagestyle{empty}
	\begin{center}
        \vspace{-4mm}
		\centerline{\includegraphics[width=0.99\linewidth]{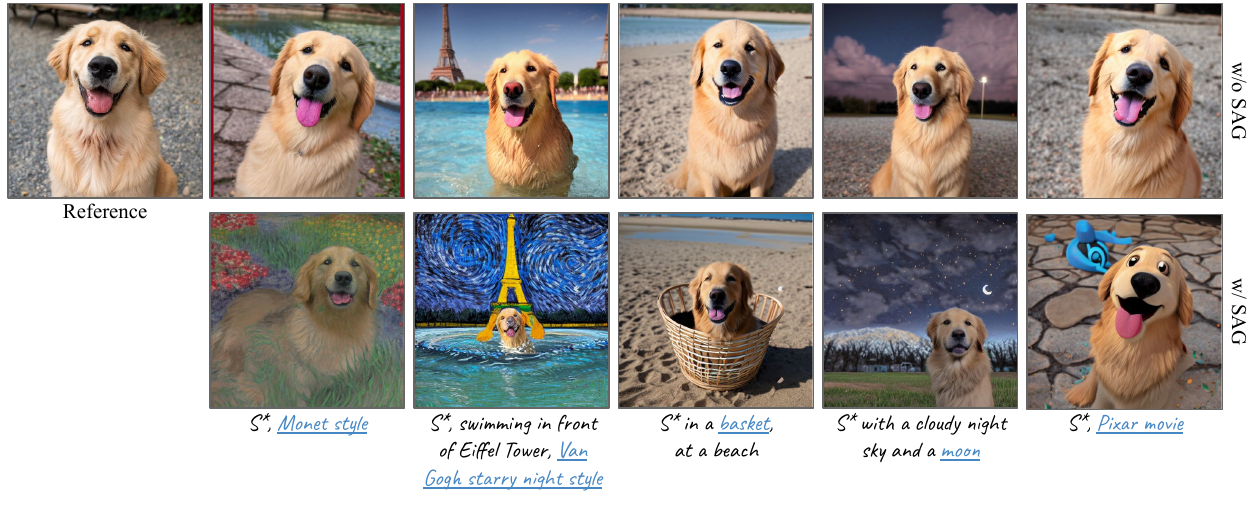}}
        \vspace{-1mm}
		\captionof{figure}{\textbf{Addressing Content Ignorance.} Given user-provided subject images, a part of the content specified in the text prompt (highlighted in \underline{{\color{cvprblue}{blue}}}) are overlooked. Our \textit{Subject-Agnostic Guidance (SAG)} aligns the output more closely with both the target subject and text prompt. Here \texttt{S$^*$} denotes a pseudo-word, with its text embedding replaced by a learnable subject embedding.}
		\label{fig:teaser}
	\end{center}
}]
\maketitle

\begin{abstract}
In subject-driven text-to-image synthesis, the synthesis process tends to be heavily influenced by the reference images provided by users, often overlooking crucial attributes detailed in the text prompt. 
In this work, we propose \textbf{Subject-Agnostic Guidance (SAG)}, a simple yet effective solution to remedy the problem. 
We show that through constructing a subject-agnostic condition and applying our proposed dual classifier-free guidance, one could obtain outputs consistent with both the given subject and input text prompts.  
We validate the efficacy of our approach through both optimization-based and encoder-based methods.
Additionally, we demonstrate its applicability in second-order customization methods, where an encoder-based model is fine-tuned with DreamBooth. 
Our approach is conceptually simple and requires only minimal code modifications, but leads to substantial quality improvements, as evidenced by our evaluations and user studies.
\end{abstract}
\section{Introduction}
Subject-driven text-to-image synthesis focuses on generating diverse image samples, conditioned on user-given text descriptions and subject images. This domain has witnessed a surge of interest and significant advancements in recent years.
Optimization-based methods~\cite{ruiz2023dream,gal2023image,sohn2023styledrop} tackle the problem by overfitting pre-trained text-to-image synthesis models~\cite{saharia2022photorealistic,rombach2022high} and text tokens to the given subject. 
Recently, encoder-based approaches~\cite{jia2023taming,chen2023subject,wei2023elite} propose to train auxiliary encoders to generate subject embeddings, bypassing the necessity of per-subject optimization.

In the aforementioned approaches, both the embeddings and networks are intentionally tailored to closely fit the target subject. As a consequence, these learnable conditions tend to dominate the synthesis process, often obscuring the attributes specified in the text prompt.
For instance, as shown in Fig.~\ref{fig:teaser}, when employing \texttt{S$^*$}\footnote{\texttt{S$^*$} denotes a pseudo-word, where its embedding is substituted by a learnable subject embedding.} alongside the style description \texttt{Monet style}, the desired style is not appropriately synthesized.
Such observations underscore that the network struggles to prioritize key content in the existence of learnable components.
To address the \textit{content ignorance} issue, existing solutions modify the training process through additional regularization~\cite{wei2023elite,ruiz2023dream}, leading to improved performance.

In this work, we present \textbf{\textit{Subject-Agnostic Guidance (SAG)}}, an approach that diverges from traditional methodologies. Our strategy emphasizes attending to subject-agnostic attributes by diminishing the influence of subject-specific attributes, accomplished using classifier-free guidance.
Differing from standard classifier-free guidance~\cite{ho2022classifier}, our method incorporates a subject-agnostic condition\footnote{The construction of this condition varies based on the specific customization approach used.}. Subsequently, our proposed \textit{Dual Classifier-Free Guidance (DCFG)} is employed to enhance attention directed towards subject-agnostic attributes.
Crucially, motivated by the observation that structures are constructed during early iterations~\cite{huang2023collaborative,choi2022perception}, we temporarily replace the subject-aware condition with a subject-agnostic condition at the beginning of the iteration process. Following the construction of coarse image structures, the original subject-aware condition is reintroduced to refine customized details.

Our SAG is elegant in both design and implementation, seamlessly blending with existing methods. We showcase the efficacy of SAG using both optimization-based and encoder-based approaches. Furthermore, we delve into its applicability in second-order customization, with an encoder-based model fine-tuned via DreamBooth~\cite{ruiz2023dream}. Qualitative and quantitative evaluations as well as user feedback verify our robustness, succinctness, and versatility.

In the evolving realm of subject-driven text-to-image synthesis, challenges have emerged due to over-tailored embeddings and networks. These often inherit crucial attributes. While existing solutions modify training to address these issues, our novel \textbf{\textit{Subject-Agnostic Guidance (SAG)}} provides a distinct approach. Seamlessly integrating with prevalent methods, SAG emphasizes a more balanced synthesis process. Its effectiveness is demonstrated through various methodologies and supported by user feedback.
\section{Related Work}
\noindent\textbf{Diffusion Model for Text-To-Image Synthesis.}
Typically, given natural language descriptions, a text encoder such as CLIP~\cite{radford2021learning} or T5~\cite{raffel2020exploring} is employed to derive the text embedding. This embedding is then fed into the diffusion model for the generation phase.
Earlier approaches~\cite{ramesh2022hierarchical} operated directly within the high-resolution image space for generation.
While these methods yielded promising outcomes, the direct iteration in high-resolution space poses significant computational challenges.
In light of these constraints, considerable efforts have been devoted to enhancing generation efficiency.
For instance, Imagen~\cite{saharia2022photorealistic} employs a multi-stage diffusion model. It starts by synthesizing a $64\times64$ resolution image based on the input text prompt and subsequently employs a series of super-resolution modules to increase the resolution to $1024\times1024$.
Benefiting from optimized architectures in the super-resolution stages, this cascaded approach considerably reduces computational overhead compared to direct high-resolution image synthesis.
Latent Diffusion~\cite{rombach2022high} transitions the generation process to a low-resolution feature space to improve efficiency. Initially, a VAE~\cite{kingma2014auto} or VQGAN~\cite{esser2021taming,van2017neural} is pre-trained. During training, images are encoded into low-resolution features using the pre-trained encoder, and the diffusion model aims to reconstruct these encoded features. In the inference stage, the trained diffusion model produces a feature which is subsequently decoded using the pre-trained module to render the final output image.

\vspace{0.1cm}
\noindent\textbf{Subject-Driven Image Synthesis.}
Subject-driven text-to-image synthesis~\cite{kumari2022multi,han2023svdiff,tewel2023keylocked,chen2023disenbooth,shi2023instantbooth,li2023blip,voynov2023p+,liu2023cones,chen2023photoverse,arar2023domain,li2024stylegan,hu2024instruct} is a sub-branch of text-to-image synthesis~\cite{saharia2022photorealistic,rombach2022high,chang2023muse,yu2022scaling,kang2023scaling,ding2022cogview2,dalle3} with an additional requirement that the primary attributes in the output aligns with the subjects provided by the user.
Existing research~\cite{ruiz2023dream,gal2023image,han2023highly,valevski2023face0} has demonstrated that subject information can be encoded as a subject-aware embedding through test-time optimization, given several reference images.
For instance, Textual Inversion~\cite{gal2023image} leverages pre-trained synthesis networks and optimizes a special token while keeping the network static.
DreamBooth~\cite{ruiz2023dream} shares a similar premise but also fine-tunes the network to enhance subject consistency.
To bypass test-time optimization, which restricts instant feedback, recent studies~\cite{jia2023taming,chen2023subject} advocate the use of an encoder to encapsulate subject information.
However, despite advancements in both quality and speed, the encoded subject information often dominates the synthesis process, resulting in inadequately capture of subject information .
In this study, we introduce \textit{Subject-Agnostic Guidance (SAG)} to rectify this challenge.
Our SAG focuses on enhancing subject-agnostic attributes, diminishing the influence of subject-specific elements through our dual classifier-free guidance.
We illustrate that SAG not only enhances consistency to the input captions but also maintains fidelity to the subject.
\section{Methodology}
In this work, we introduce an intuitive and effective method to enhance content alignment. We first provide the background for our approach, followed by the discussion of our method -- \textit{Subject-Agnostic Guidance}.
\begin{figure}[!t]
  \includegraphics[width=0.46\textwidth]{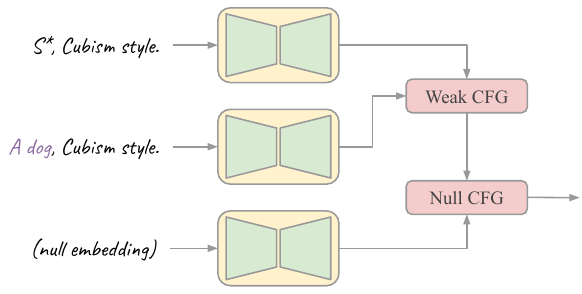}
  \caption{\textbf{Overview of SAG.} Given a subject-aware embedding, we first construct a subject-agnostic embedding. These embeddings are subsequently used in our dual classifier-free guidance (DCFG), which consists of weak classifier-free guidance and null-classifier-free guidance. Null CFG adopts a constant weight (Eqn.~\ref{eq:cfg}) and Weak CFG adopts a variable weight (Eqn.~\ref{eq:DCFG}).}
  \label{fig:overview}
\end{figure}
\subsection{Preliminaries}
\subsubsection{Diffusion Model}
The diffusion process transforms a data distribution to a Gaussian noise distribution by iteratively adding noise. 
Diffusion model is a class of generative models that invert the diffusion process through iterative denoising. 
Extended from the original unconditional model~\cite{ho2020denoising}, recent works demonstrate huge success by conditioning diffusion models on various modalities, including text~\cite{huang2023reversion,sheynin2023knn,chefer2023attend}, segmentation~\cite{huang2023collaborative,bar2023multidiffusion,mou2023t2i}, and many more~\cite{wang2023exploiting,su2023identity,li2023gligen}.

Let $\mathbf{x}_0$ be the input image, and $\mathbf{c}$ be the condition. 
During training, a noisy image $\mathbf{x}_t$ is obtained by adding Gaussian noise $\bm{\epsilon}_t$ to $\mathbf{x}_0$. 
The network is trained to predict the added noise, given the noisy image and condition as input. It is generally optimized with a single denoising objective: 
\begin{equation}
\label{eq:objective}
    \mathcal{L}_d = ||\bm{\epsilon}(\mathbf{x}_t, \mathbf{c}) - \bm{\epsilon}_t||_2^2,
\end{equation}
where $\bm{\epsilon}_t$ the noise added to the input image, and $\bm{\epsilon}(\mathbf{x}_t, \mathbf{c})$ corresponds to the noise estimated by the network. 
Here $\mathbf{x}_t$ and $\mathbf{c}$ refer to the noisy image and condition, respectively.
During inference, the process starts with a pure Gaussian noise $\mathbf{x}_{T_0}$, and the trained network is iteratively applied to obtain a series of intermediate outputs $\{\mathbf{x}_{T_0-1}, \mathbf{x}_{T_0-2}, \cdots, \mathbf{x}_{0}\}$, where $\mathbf{x}_{0}$ is the final output.  

\subsubsection{Classifier-Free Guidance}
Similar to classifier guidance~\cite{dhariwal2021diffusion}, classifier-free guidance is designed to trade between image quality and diversity, but without the need of a classifier. It is widely adopted in existing works~\cite{villegas2023phenaki,zhang2023adding}. 

During training, an unconditional diffusion model is jointly trained by randomly replacing the input condition $\mathbf{c}$ by a null condition $\bm{\phi}$. Once trained, during each iteration $t$, a weighted sum of the conditional output and the unconditional output is computed:
\begin{equation}
\label{eq:cfg}
    \bm{\tilde{\epsilon}}_t = (1 + w)\cdot\bm{\epsilon}(\mathbf{x}_t, \mathbf{c}) - w\cdot\bm{\epsilon}(\mathbf{x}_t, \bm{\phi}).
\end{equation}
In general, a larger $w$ produces better quality, whereas a smaller $w$ yields greater diversity. 

\subsection{Subject-Agnostic Guidance}
\label{subsec:SAG}
In this section, we introduce the concept of \textbf{\textit{Subject-Agnostic Guidance (SAG)}}. The essence of SAG is anchored in formulating a \textit{subject-agnostic embedding} based on the inputs provided by users. The embedding is then used in our \textit{dual classifier-free guidance (DCFG)} in generating outputs that align with both the subject and text prompt. We delve into the details of constructing subject-agnostic embeddings in Sec.~\ref{subsubsec:CAE}, and discuss our dual classifier-free guidance in Sec.~\ref{subsubsec:DCFG}.

\subsubsection{Subject-Agnostic Embeddings}
\label{subsubsec:CAE}
The construction of subject-agnostic embeddings depends on the choice of methods. Existing approaches generally fall into two categories: \textbf{Learnable Text Token} and \textbf{Separate Subject Embedding}. In this section, we discuss the construction of subject-agnostic embeddings in these two approaches.

\vspace{1mm}
\noindent\textbf{Learnable Text Token.}
Given images of a reference subject, the learnable text token approach derives a token embedding that captures the identity of the subject, either through fine-tuning~\cite{gal2023image,voynov2023p+} or by using an encoder~\cite{wei2023elite,arar2023domain}. The resultant token embedding, combined with the token embedding of the text description, is processed by text encoders such as CLIP~\cite{radford2021learning} and T5~\cite{raffel2020exploring} to produce a subject-aware embedding.

To construct a subject-agnostic embedding, we replace the derived token embedding with one from a general description of the subject. This strategy ensures that the synthesis process is not dominated by any adaptable components, thereby allowing the model to focus attention on the attributes specified in the text prompt.

Let $\mathbf{c}$ be the text condition containing the learnable token \texttt{S$^*$}. %
We define a subject-agnostic condition $\mathbf{c}_0$ by replacing the token \texttt{S$^*$} by a generic descriptor. For example, assuming the target subject is a dog and 
\begin{center}
    \mbox{$\mathbf{c}=$ \texttt{A pencil sketch of S$^*$}}
\end{center}
we construct $\mathbf{c}_0$ as
\begin{center}
    $\mathbf{c}_0=$ \texttt{A pencil sketch of a dog}
\end{center}
The generic descriptor is chosen as a noun describing the subject. 

\vspace{1mm}
\noindent\textbf{Separate Subject Embedding.} 
Instead of encoding the subject identity to a learnable text token, the separate subject embedding approach~\cite{jia2023taming,chen2023subject} adopts an independent embedding. This embedding is then integrated into the network via auxiliary operations. For instance, Jia \etal~\cite{jia2023taming} employ the CLIP image encoder to encapsulate the subject information into an embedding, which is then injected to Imagen~\cite{saharia2022photorealistic} using cross attention.

To construct the subject-agnostic embedding, we opt for a direct method -- setting both the subject embedding and its corresponding attention mask to zero. This disables attention to the subject, directing focus towards subject-agnostic information.

\subsubsection{Dual Classifier-Free Guidance}
\label{subsubsec:DCFG}
In this section, we introduce the \textit{Dual Classifier-Free Guidance (DCFG)}, designed primarily to address the issue of content ignorance by attenuating the subject-aware condition. Our DCFG requires no modifications of the training process. It simply requires the application of an additional classifier-free guidance using the subject-aware condition $\mathbf{c}$ and the subject-agnostic condition $\mathbf{c}_0$. The derived feature is subsequently merged with the null condition $\bm{\phi}$ within a conventional classifier-free guidance.

\vspace{1mm}
\noindent\textbf{Weak Classifier-Free Guidance.}
Given the subject-aware condition $\mathbf{c}$ and the subject-agnostic condition $\mathbf{c}_0$, we first perform classifier-free guidance using $\mathbf{c}$ and $\mathbf{c}_0$. Incorporating $\mathbf{c}_0$ into the synthesis process directs the generation towards subject-agnostic content, representing a weaker version of the desired generation.
When subject information is absent, the model more effectively creates the correct outline and structure, generating outputs that align with both the subject and text description.

Differing from the conventional classifier-free guidance, where the guidance weight $w$ often remains constant during the denoising process, we implement a time-varying scheme to enhance performance.
Building on the observation that earlier iterations emphasize structure construction~\cite{huang2023collaborative,choi2022perception}, we highlight the subject-agnostic condition during the initial phases. Specifically, we adopt a time-varying weighting strategy, suppressing subject information in the early stages:
\begin{equation}
\label{eq:DCFG}
    \bm{\bar{\epsilon}}_t = (1 + w_t)\cdot\bm{\epsilon}(\mathbf{x}_t, \mathbf{c}) - w_t\cdot\bm{\epsilon}(\mathbf{x}_t, \mathbf{c_0}),
\end{equation}
where $w_t$ denotes the guidance weight, similar to $w$ in Eqn.~\ref{eq:cfg}. Since a larger $w_t$ corresponds to a larger contribution from $\mathbf{c}$, $w_t$ is devised as a \textit{non-increasing} function with respect to the iteration $t$.
In this work, we find that a simple piecewise constant scheme suffices to produce promising results:
\begin{equation}
\label{eq:DCFG_w}
    w_t = \begin{cases}
    r &\text{if } 0 \leq t \leq T, \\
    -1\quad &\text{if } T < t \leq 1.\\
    \end{cases}
\end{equation}
Here $0 \leq T \leq 1$ and $r \geq -1$ are pre-determined constants, which will be ablated in Sec.~\ref{sec:ablation}. Essentially, in the early stages (\ie, when $t\approx1$), we use solely the subject-agnostic condition to establish the structure and outline of the output. The subject information is integrated in the subsequent stages.

\vspace{0.1cm}
\noindent\textbf{Null Classifier-Free Guidance.}
The null classifier-free guidance is identical to the conventional classifier-free guidance, leveraging the null condition to encourage diversity. We adopt a constant guidance weight throughout iterations. Specifically, the output $\bm{\bar{\epsilon}}_t$ of the weak-classifier-free guidance is used in place of $\bm{\epsilon}(\mathbf{x}_t, \mathbf{c})$ in the conventional classifier-free guidance (Eqn.~\ref{eq:cfg}):
\begin{equation}
\label{eq:DCFG_null}
    \bm{\tilde{\epsilon}}_t = (1 + w)\cdot\bm{\bar{\epsilon}}_t - w\cdot\bm{\epsilon}(\mathbf{x}_t, \bm{\phi}).
\end{equation}

\begin{figure*}
  \includegraphics[width=0.99\textwidth]{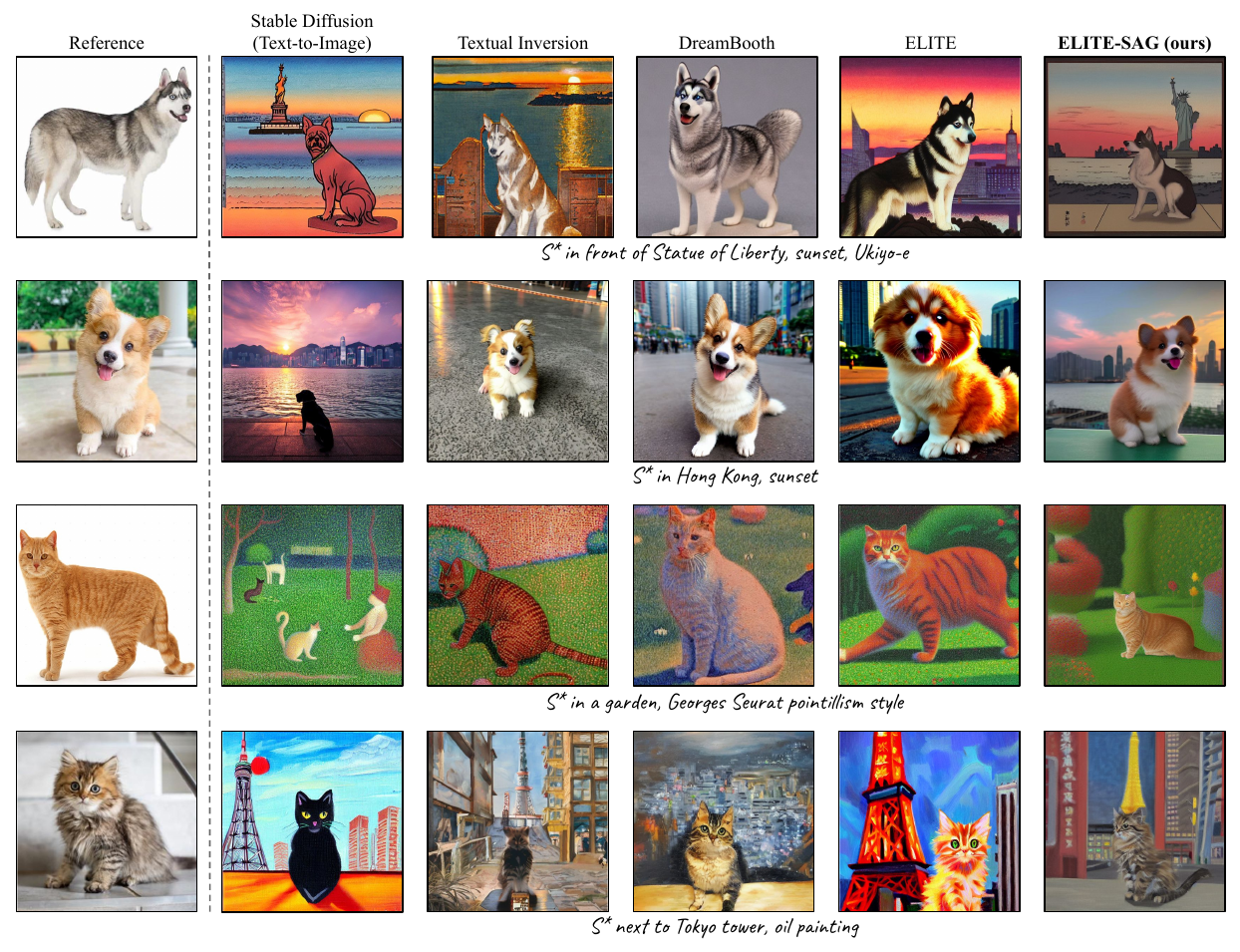}
  \caption{\textbf{SAG on ELITE~\cite{wei2023elite}.} Our ELITE-SAG produces outputs that are more faithful to text prompts while still preserving subject identity. For Stable Diffusion, we generate pure text-to-image results by substituting ``\texttt{S}$^*$'' with ``\texttt{A dog}'' or ``\texttt{A cat}''.}
  \label{fig:comparison}
\end{figure*}

\section{Experiments}
\label{sec:exp}
To validate the efficacy of SAG, we conduct experiments across multiple approaches, namely Textual Inversion~\cite{gal2023image} (optimization-based), ELITE~\cite{wei2023elite} (encoder-based), SuTI~\cite{chen2023subject} (encoder-based), and DreamSuTI~\cite{chen2023subject} (second-order).

\subsection{ELITE}
First, we examine the performance improvement when applying SAG to ELITE~\cite{wei2023elite}. In this study, we simplify its architecture by using only the global mapping branch. The settings are as follows:

\vspace{1mm}
\noindent\textbf{Training.}
To promote the learning of subject information, we create a domain-specific (\eg, animals) text-image dataset where the text caption incorporates the specialized token. Specifically, we gather images from a pre-defined category and employ straightforward templates such as \mbox{\texttt{A photo of S}$^*$} for the corresponding captions. During training, the token corresponding to \texttt{S}$^*$ is substituted with the output of the encoder.
The condition is subsequently fed into the text encoder.

As discussed in concurrent work~\cite{jia2023taming}, text prompts generated using templates and captioning models~\cite{chen2022pali} have inherent limits to their diversity. Moreover, training within narrow domains may harm generation diversity. To counteract this, we employ a general-domain dataset containing detailed text descriptions for regularization.
Training on a broad array of text captions ensures the model retains its text-understanding abilities.

During the training phase, the domain-specific and general-domain datasets are sampled with probabilities \mbox{$p\leq1$} and $(1 - p)$, respectively.
Given that the general-domain dataset serves primarily for regularization, we allocate a higher value to $p$, greater than 0.5, emphasizing subject encoding.
%

Since the subject-agnostic condition $\mathbf{c}_0$ is also natural language, no modification to the original denoising objective (Eqn.~\ref{eq:objective}) is needed. 
Additionally, we adopt a regularization to the learnable token~\cite{wei2023elite} by constraining its $\ell_2$-norm.
The effective training loss is:
\begin{equation}
    \mathcal{L} = \mathcal{L}_{d} + ||s||^2\,,
\end{equation}
where $s$ denotes the output of the subject encoder. The remaining part of the training is identical to the training of conventional text-to-image networks. 

\begin{figure*}
  \includegraphics[width=0.99\textwidth]{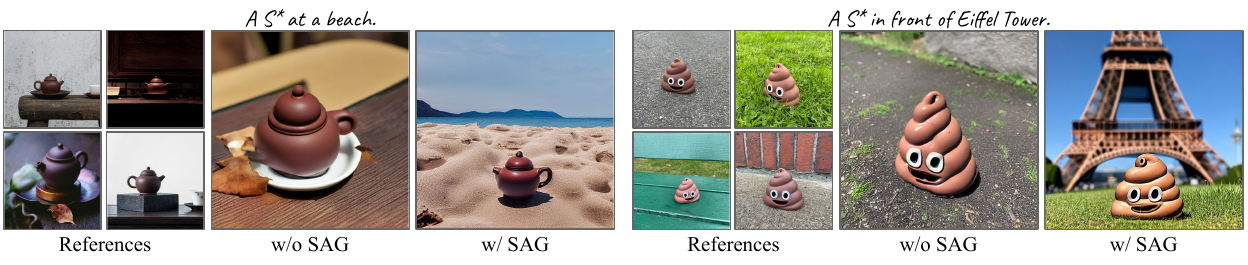}
  \caption{\textbf{SAG on Textual Inversion~\cite{gal2023image}.} Our SAG improves text alignment without sacrificing the identity of the subject.}
  \label{fig:ti_comparison}
\end{figure*}

\vspace{0.1cm}
\noindent\textbf{Inference.}
For each input image, we use the encoder to map the target subject into a text token. This learnable token is then combined with the text description to form the input condition $\mathbf{c}$. The subject-agnostic condition is $\mathbf{c}_0$ is then constructed following the process discussed in Sec.~\ref{subsubsec:CAE}.  
Starting from random Gaussian noise $\mathbf{x}_T$, the fine-tuned network iteratively denoises the intermediate outputs. 
Instead of applying the conventional classifier-free guidance, our SAG is employed. 

\vspace{1mm}
\noindent\textbf{Implementation.}
We adopt the pre-trained Stable Diffusion~\cite{rombach2022high} as the synthesis network, which uses CLIP~\cite{radford2021learning} as the text encoder.
For the subject encoder, we use the CLIP image encoder and a three-layer MLP to obtain the learnable token. 
During training, only the cross-attention layers in Stable Diffusion and the MLP are trained, all other weights are being fixed.

We use an internal text-image dataset for training. To construct the domain-specific dataset, we extract images containing \textit{dogs} and \textit{cats} from the meta-dataset. The remaining part is used as our general-domain dataset. 
The dataset mixing ratio is $0.1$. 
The proposed method is implemented in JAX~\cite{jax2018github}. 
The detailed experimental settings will be discussed in the supplementary material.

\begin{table}[!t]
    \centering
    \caption{\textbf{Quantitative Comparison}. Our ELITE-SAG yields improved performance in both text and subject alignment.}
    \scalebox{0.9}{
    \begin{tabular}{l|c|c|c}
        Methods               & CLIP-T~$\uparrow$ & CLIP-I~$\uparrow$ & DINO~$\uparrow$ \\
        \hline
        DreamBooth~\cite{ruiz2023dream}    & 0.315                     & 0.785                     & 0.651                    \\
        Textual Inversion~\cite{gal2023image} & 0.339                 & 0.751                     & 0.571                    \\
        ELITE~\cite{wei2023elite}         & 0.342                     & 0.751                     & 0.586                    \\
        ELITE-SAG \textbf{(ours)}          & \textbf{0.344}            & \textbf{0.790}            & \textbf{0.671}           \\
    \end{tabular}

    }
    \label{tab:quan_comp}
\end{table}

\begin{table}[!t]
    \centering
    \caption{\textbf{User Study}. Across all three compared methods, the majority of raters favor the results produced by our approach.}
    \scalebox{0.9}{
    \begin{tabular}{l|c|c|c}
        \% Prefer Ours           & Subject Align. & Text Align.  & Quality \\
        \hline
        DreamBooth~\cite{ruiz2023dream}       & 52\%    & 68\%  & 60\%    \\
        Textual Inversion~\cite{gal2023image} & 64\%    & 76\%  & 84\%    \\
        ELITE~\cite{wei2023elite}             & 56\%    & 80\%  & 76\%    \\
    \end{tabular}
    }
    \label{tab:user_study}
\end{table}

\vspace{1mm}
\noindent\textbf{Comparison.}
We compare our modified model, \textit{ELITE-SAG}, with three existing works: DreamBooth~\cite{ruiz2023dream}, Textual Inversion~\cite{gal2023image}, and ELITE~\cite{wei2023elite}. In this section, we assume the existence of only one reference image. As illustrated in Fig.~\ref{fig:comparison}, while Stable Diffusion exhibits high text alignment, the compared methods often fall short in generating results faithful to text prompts in the presence of additional subject images. In contrast, with our SAG, outputs adhering to both text captions and reference subjects are consistently generated.

We also conduct a quantitative comparison as presented in Table~\ref{tab:quan_comp}, utilizing CLIP~\cite{radford2021learning} and DINO~\cite{caron2021emerging} scores. Specifically, the image feature similarities of CLIP~\cite{radford2021learning} and DINO underscore that SAG enhances subject fidelity, while the text feature similarity indicates that SAG improves text alignment.
Furthermore, our user study depicted in Table~\ref{tab:user_study} reveals that more than half of the raters prefer our method when compared to the aforementioned methods, thereby corroborating the effectiveness of SAG.

\subsection{Textual Inversion}
Textual Inversion~\cite{gal2023image} is an optimization-based method for customization. For each given subject, Textual Inversion learns a text token to represent the subject. As discussed in Sec.~\ref{subsubsec:CAE}, the subject-agnostic embedding is generated by replacing the learned special token by a generic description. Then, the conventional CFG is replaced by our SAG. The remaining generation pipeline remains unchanged.

As illustrated in Fig.~\ref{fig:ti_comparison}, the absence of SAG leads to generation dominated by the optimized text token, resulting in suboptimal text alignment. Conversely, the incorporation of SAG enables the model to produce outputs that align more closely with the text description, while preserving the identity of the subject.

\begin{figure*}
  \includegraphics[width=\textwidth]{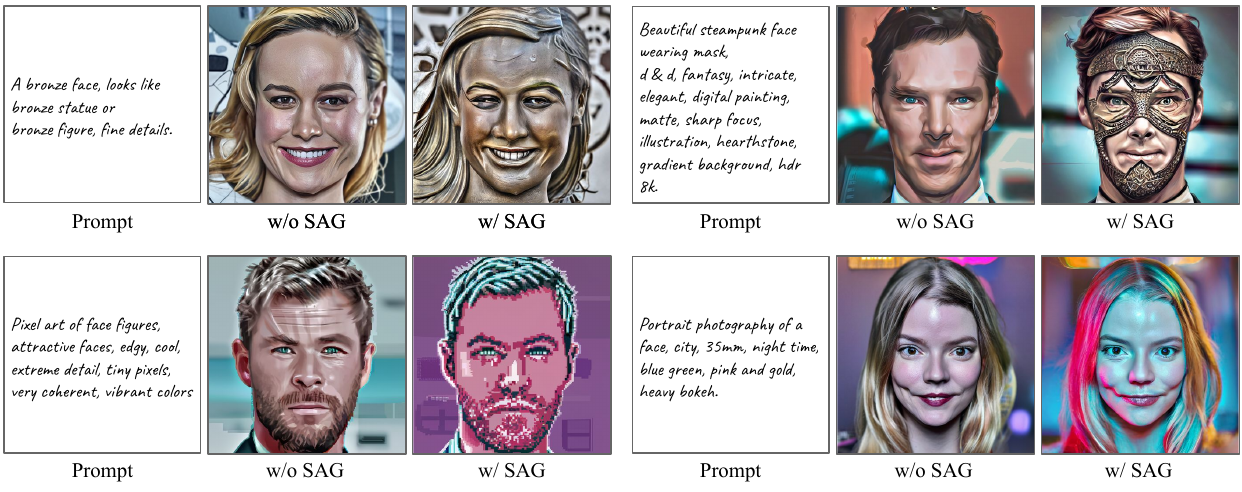}
  \caption{\textbf{SAG on SuTI~\cite{chen2023subject}.} When applying SAG on SuTI, the subject is discarded during initial iterations, yielding outputs with markedly improved text alignment. Reference images are not provided to protect privacy.}
  \label{fig:suti_comparison}
  \vspace{2mm}
\end{figure*}

\begin{figure*}
  \includegraphics[width=\textwidth]{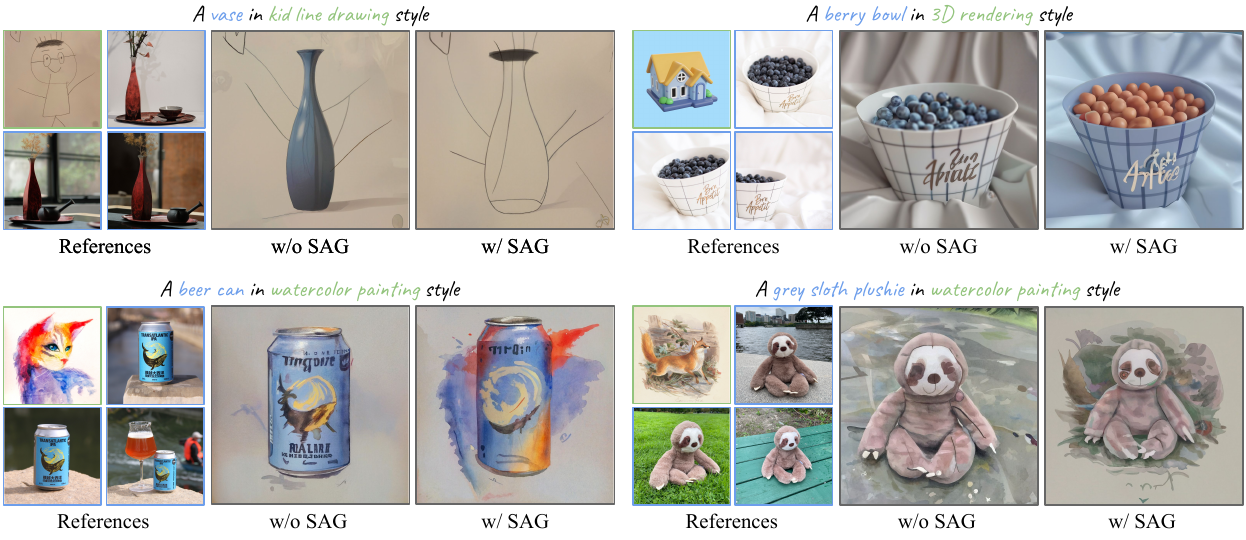}
  \caption{\textbf{SAG on DreamSuTI~\cite{chen2023subject}.} Even after fine-tuning with DreamBooth to adapt to the specified style, the generated results tend to be dominated by the subjects, leading to an inadequate style-alignment. Our SAG addresses this issue by diminishing the influence of subjects, thereby ensuring outputs that are well-aligned with both the text, subject, and style.}
  \label{fig:dreamsuti_comparison}
\end{figure*}

\subsection{SuTI}
Unlike ELITE, which encodes subject information into a text token, SuTI~\cite{chen2023subject} employs an encoder-based approach that leverages a distinct subject embedding. This embedding is then fed to the generation network through independent cross-attention layers. As discussed in Sec.~\ref{subsubsec:CAE}, the subject-agnostic condition, denoted as $\mathbf{c}_0$, is simply constructed by setting the subject embedding to zero.

As illustrated in Fig.~\ref{fig:suti_comparison}, without SAG, the model successfully preserves the identity of the individual provided in the reference images, yet the text alignment is inadequate. Specifically, the styles are unsatisfactory across all outputs. In contrast, employing SAG and suppressing the subject information during initial iterations significantly enhances text alignment. Consequently, the outputs exhibit both high identity preservation and improved text alignment.

\subsection{DreamSuTI}
DreamSuTI~\cite{chen2023subject} is a second-order method that fine-tunes SuTI using DreamBooth~\cite{ruiz2023dream} for compositional customization. In this section, we fine-tune SuTI with a provided style image to achieve simultaneous customization of style and subject. The subject-agnostic embedding is generated using the same method as in SuTI.

As depicted in Fig.~\ref{fig:dreamsuti_comparison}, in the presence of subject images, the outputs are dominated by the subject, resulting in a lack of style fidelity. In contrast, when applying SAG, the subject is suppressed during the early stages of generation, effectively leading to enhanced style generation.

\begin{figure*}[!t]
  \includegraphics[width=0.99\textwidth]{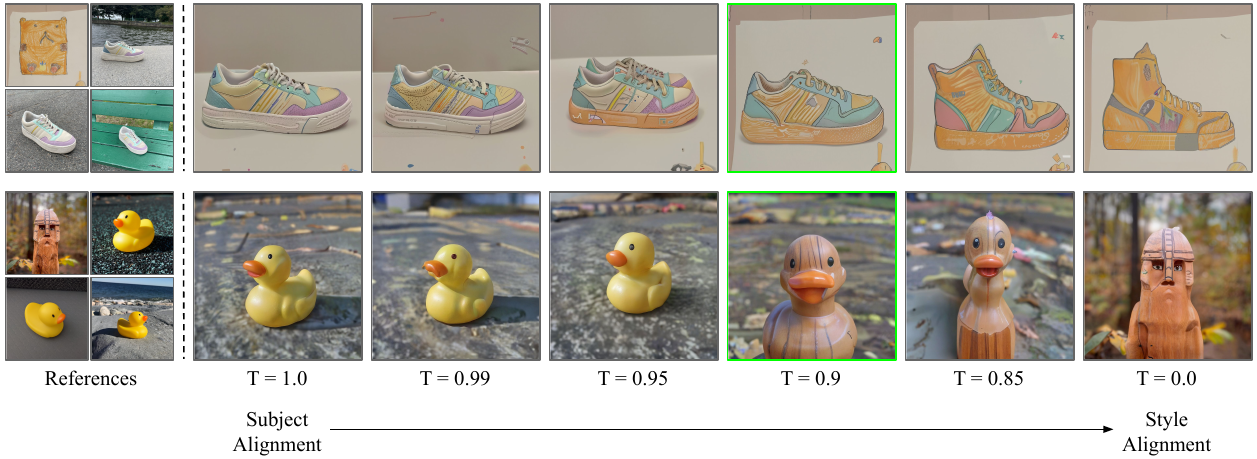}
  \caption{\textbf{Guidance Timing.} As an example, when fine-tuning SuTI~\cite{chen2023subject} to a given style using DreamBooth~\cite{ruiz2023dream}, our SAG facilitates a transition from subject-centric alignment to style-centric alignment. Here $r = 0$ is used.}
  \label{fig:guidance_T}
\end{figure*}

\begin{figure*}[!t]
  \includegraphics[width=0.99\textwidth]{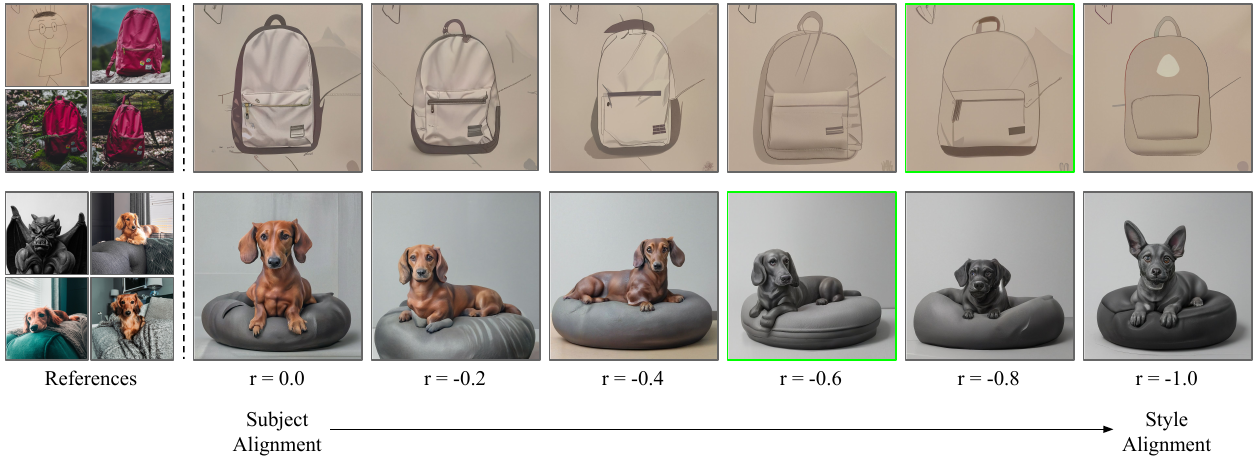}
  \caption{\textbf{Guidance Weight.} The guidance weight $r$ can be leveraged to enhance content faithfulness further. For instance, lowering $r$ results in improved style alignment in DreamSuTI. Here $T=0.9$ is used.}
  \vspace{-4mm}
  \label{fig:guidance_r}
\end{figure*}
\section{Ablations}
\label{sec:ablation}
\noindent\textbf{Guidance Timing.}
The hyper-parameter $T$ plays an important role in controlling the contribution of the subject embedding. An illustration employing DreamSuTI is provided in Fig.~\ref{fig:guidance_T}. With $r = 0$, adopting a smaller $T$ results in a stronger suppression of the subject embedding, thereby promoting a better text-alignment (\ie, style-alignment in this example). A gradual increment in $T$ facilitates a transition from style alignment to subject alignment.

\vspace{1mm}
\noindent\textbf{Guidance Weight.}
While a default value of $r=0$ (\ie, employing only the subject-aware condition in later iterations) performs well generally, decreasing $r$ facilitates the utilization of the subject-agnostic condition in subsequent iterations, thereby further enhancing content faithfulness. As depicted in Fig.~\ref{fig:guidance_r}, the inclusion of subject-agnostic conditions significantly improves the style alignment of DreamSuTI. Since no re-training is required, the values of $T$ and $r$ can be dynamically adjusted based on user preference.

\section{Limitation and Societal Impact}
\noindent\textbf{Limitation.}
While our SAG significantly enhances content alignment compared to existing methods, the quality of outputs is inherently constrained by the underlying generation model. Hence, it may still exhibit suboptimal performance for uncommon content that challenges the generation model.
However, this limitation can be mitigated by incorporating a more robust synthesis network, a direction we aim to explore in our future work.

\noindent\textbf{Societal Impact.}
This project targets at improving content alignment in customized synthesis, which holds the potential for misuse by malicious entities aiming to mislead the public. Future investigations in this domain should duly consider these ethical implications. Moreover, ensuing efforts to develop mechanisms for detecting images generated by such models emerge as a critical avenue to foster the safe advancement of generative models.
\section{Conclusion}
Subject-driven text-to-image synthesis has witnessed notable progress in recent years. However, overcoming the problem of content ignorance remains a significant challenge.
As shown in this work, this problem significantly limits the diversity of the generation.
Rather than introducing complex modules, we propose a straightforward yet effective method to address this issue.
Our Subject-Agnostic Guidance demonstrates how a balance between content consistency and subject fidelity can be achieved using a subject-agnostic condition.
The proposed method enables users to generate customized and diverse scenes without modifying the training process, making it adaptable across various existing approaches.

{
    \small
    \bibliographystyle{ieeenat_fullname}
    \bibliography{citations}
}

\clearpage

\appendix
\section{Experimental Settings}
To train ELITE-SAG, we use a subset of the the WebLI \cite{chen2022pali} dataset for training. We randomly select 70M data from the master dataset. 
We further extract 1M data containing \textit{dogs} and \textit{cats} with their face size greater than $128\times128$ as our domain-specific dataset. The remaining data is used as our general-domain dataset. 
The dataset mixing ratio is $0.1$. In this work, the weak condition $\mathbf{c_0}$ is obtained simply by replacing the special token with the class of the subject (\textit{e.g.}, ``dog'' or ``cat''). 
We train our models with 8 TPUv4 chips for 300,000 iterations. The learning rate is set to $10^{-4}$. 
The method is implemented in JAX \cite{jax2018github}. 

\begin{figure*}[!t]
    \centering
    \begin{tabular}{c}
         \includegraphics[width=0.99\textwidth]{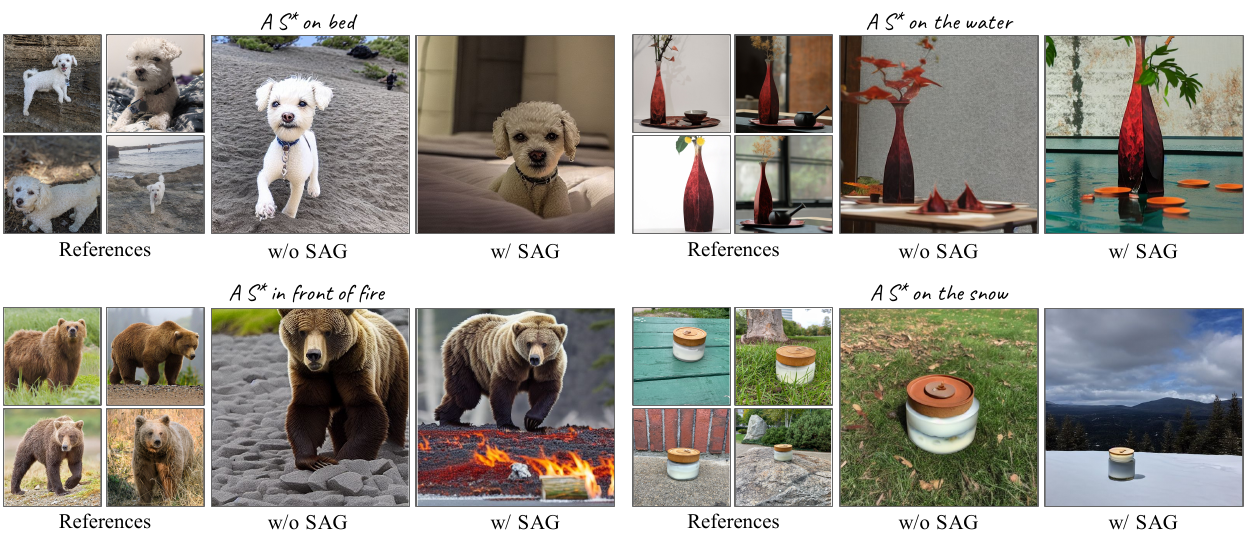}
    \end{tabular}
  \caption{\textbf{More Results on Textual Inversion.} With SAG, Textual Inversion is able to produce results that are better align with the text descriptions.}
    \label{fig:supp_ti}
\end{figure*}

\begin{figure*}[!t]
  \includegraphics[width=0.99\textwidth]{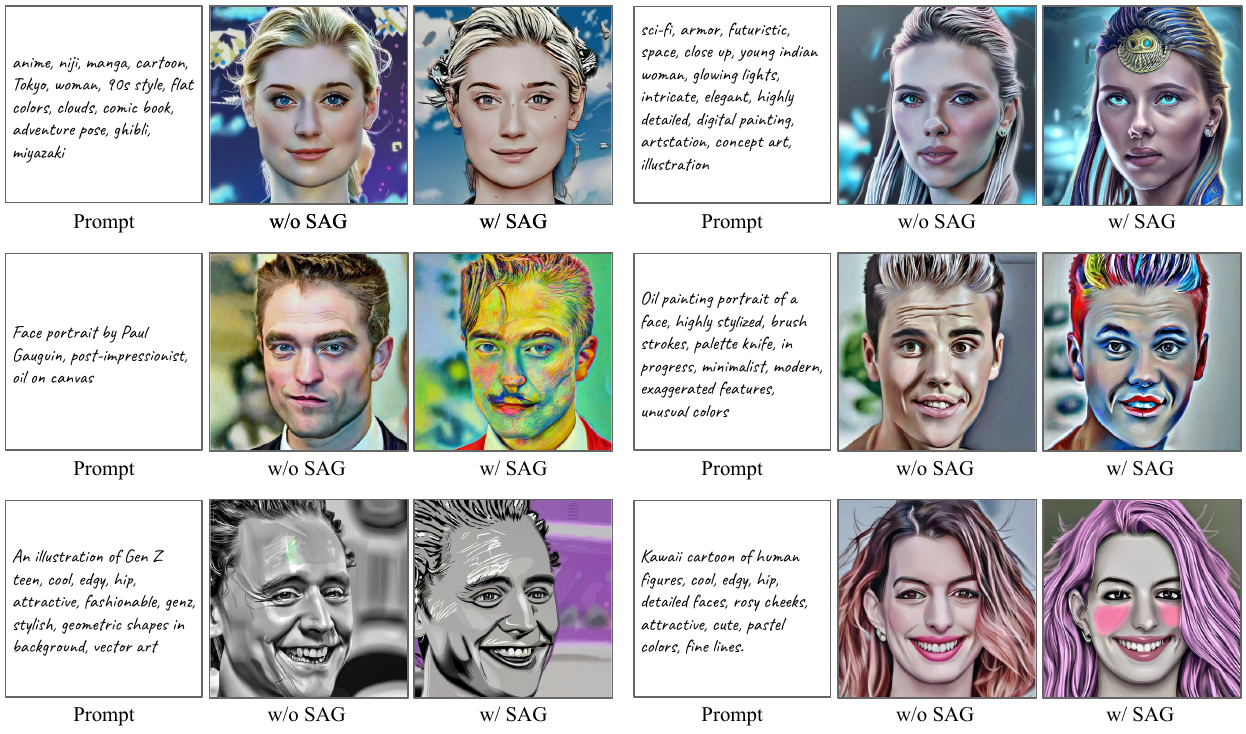}
  \caption{\textbf{More Results on SuTI.} Without SAG, while identity is successfully preserved, the outputs often fail to capture the details specified by the text prompts. In contrast, text alignment is much better with SAG, without sacrificing identity. Reference images are not provided to protect privacy.}
  \label{fig:supp_suti}
\end{figure*}

\begin{figure*}[!t]
  \includegraphics[width=0.99\textwidth]{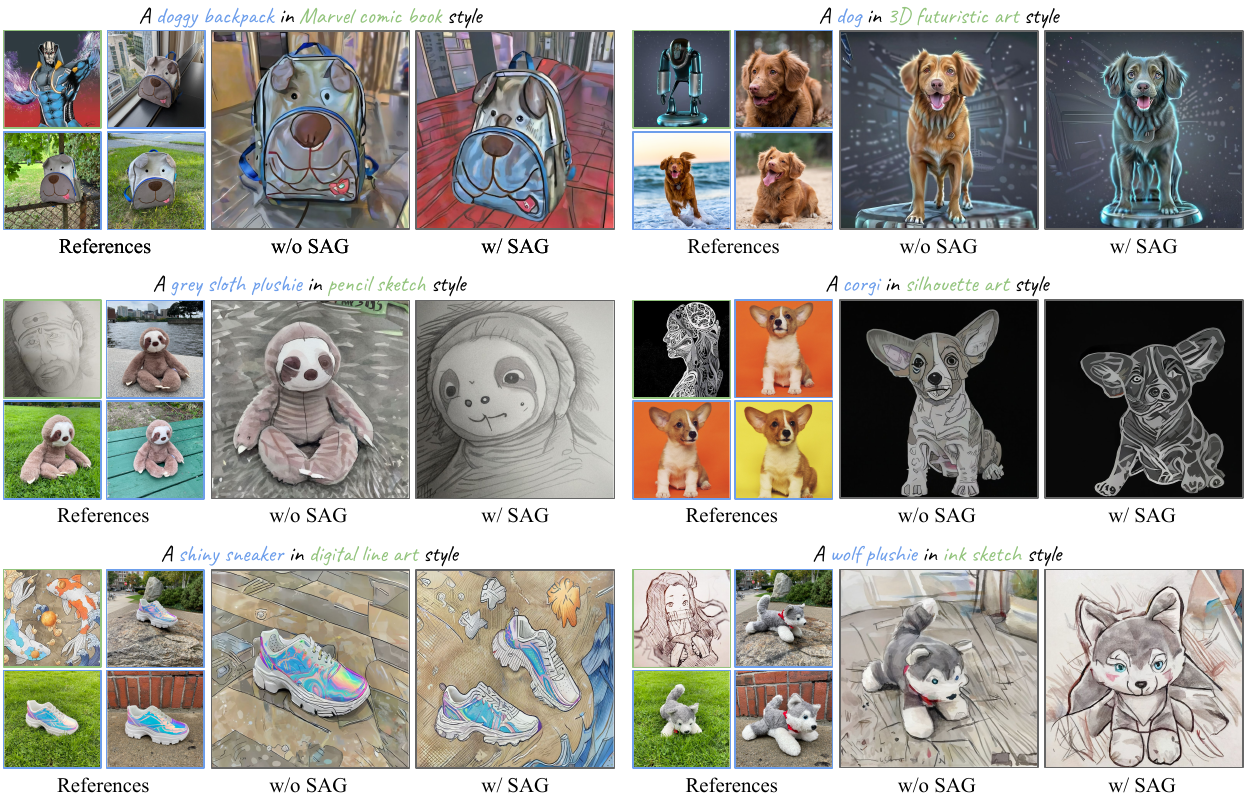}
  \caption{\textbf{More Results on DreamSuTI.} Given a SuTI fine-tuned on a given style, our SAG leads to better style alignment while being able to preserve subject identity.}
  \label{fig:supp_dreamsuti}
\end{figure*}

\begin{figure*}[!t]
  \includegraphics[width=0.9\textwidth]{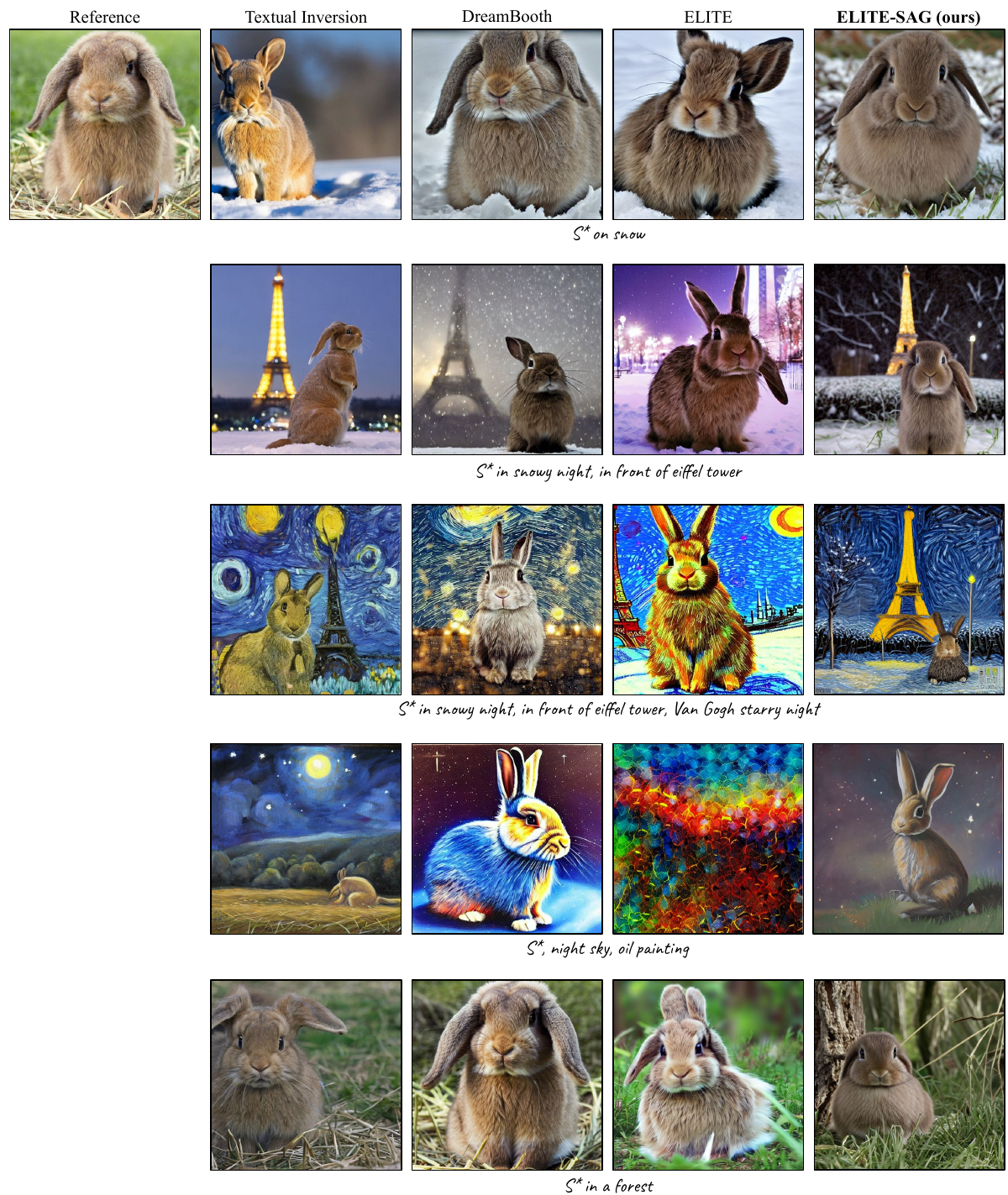}
  \caption{\textbf{More Comparison Using ELITE-SAG.} While existing works generally produce reasonable results, they often experience content ignorance or insufficient subject fidelity. This observation is especially obvious in complicated prompts, where the model has to comprehend complex relation between subjects. In contrast, ELITE-SAG achieves a balance between prompt consistency and subject fidelity.}
  \label{fig:supp_more_comparison_rabbit}
\end{figure*}

\begin{figure*}[!t]
  \includegraphics[width=0.9\textwidth]{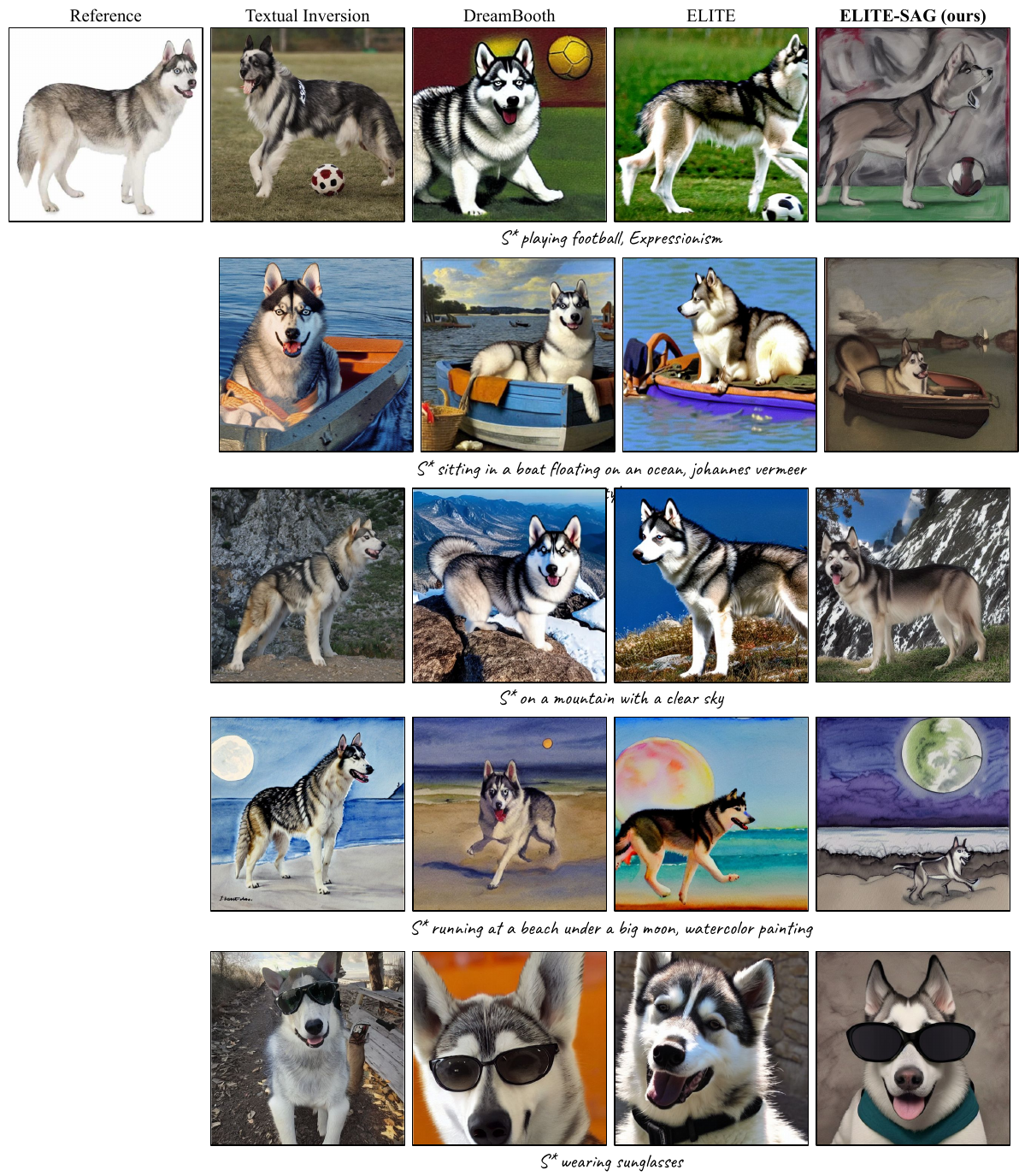}
  \caption{\textbf{More Comparison Using ELITE-SAG.} While existing works generally produce reasonable results, they often experience content ignorance or insufficient subject fidelity. This observation is especially obvious in complicated prompts, where the model has to comprehend complex relation between subjects. In contrast, ELITE-SAG achieves a balance between prompt consistency and subject fidelity.}
  \label{fig:supp_more_comparison_dog_2}
\end{figure*}

\begin{figure*}[!t]
  \includegraphics[width=0.9\textwidth]{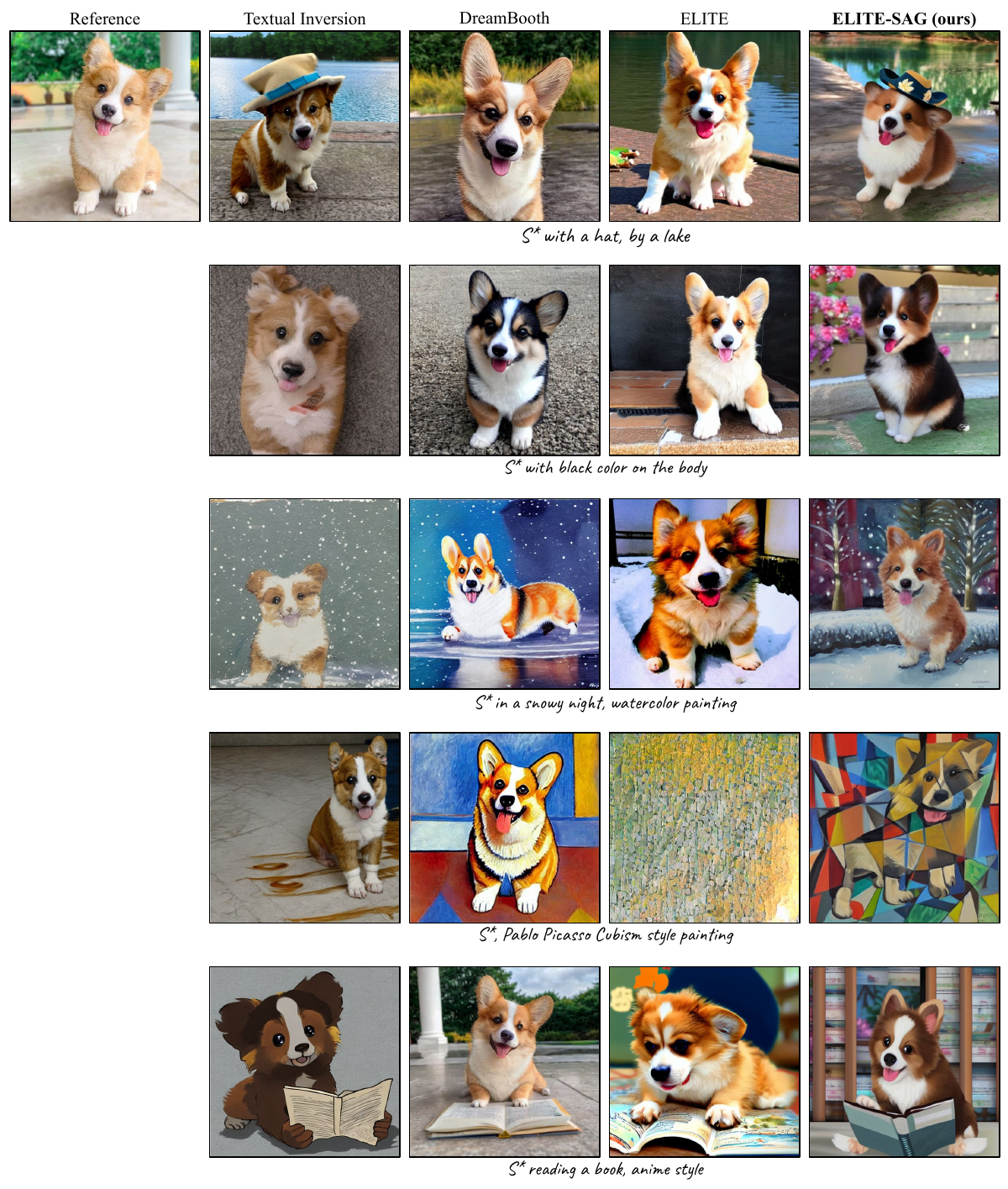}
  \caption{\textbf{More Comparison Using ELITE-SAG.} While existing works generally produce reasonable results, they often experience content ignorance or insufficient subject fidelity. This observation is especially obvious in complicated prompts, where the model has to comprehend complex relation between subjects. In contrast, ELITE-SAG achieves a balance between prompt consistency and subject fidelity.}
  \label{fig:supp_more_comparison_dog_1}
\end{figure*}

\begin{figure*}[!t]
  \includegraphics[width=0.9\textwidth]{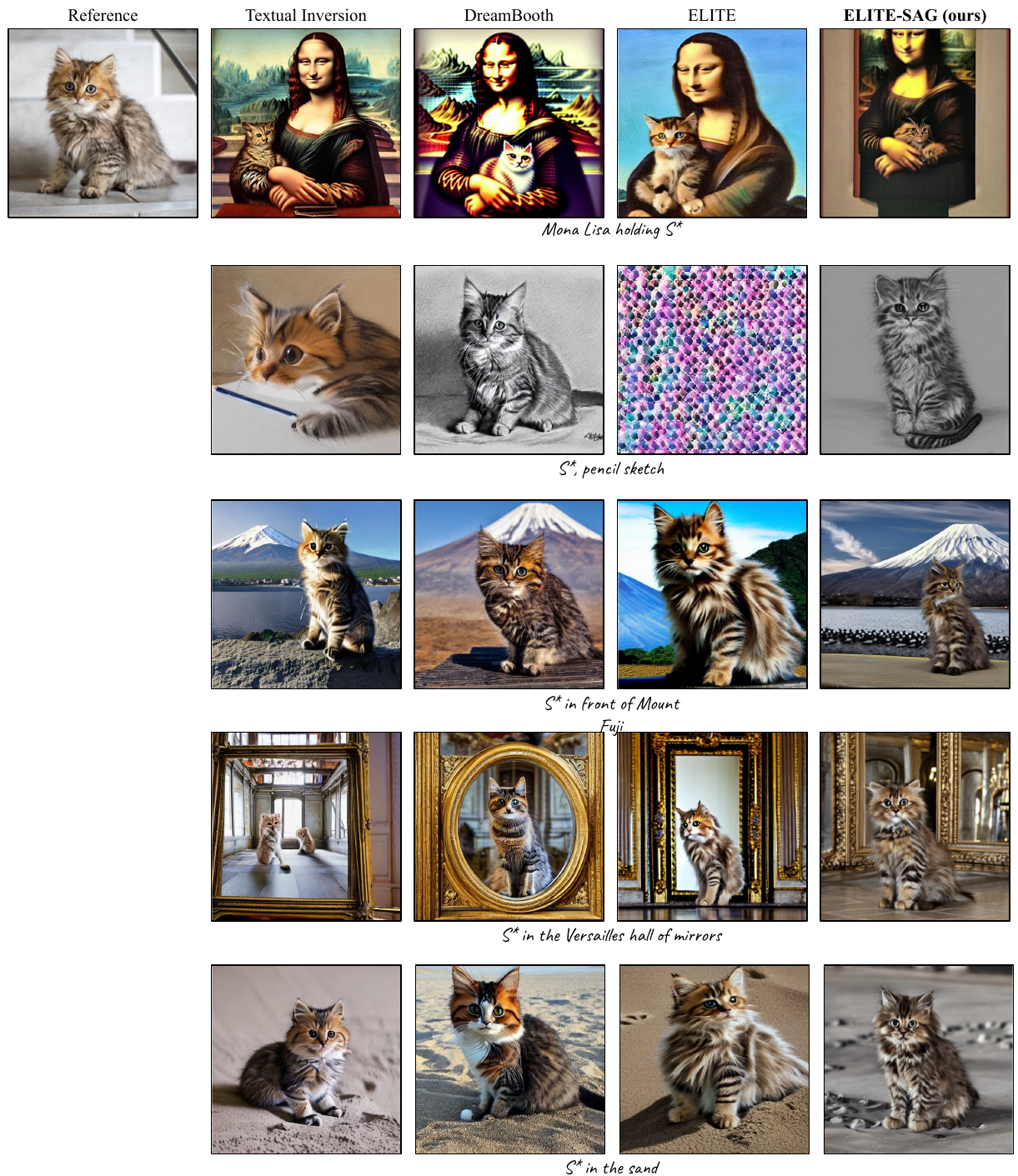}
  \caption{\textbf{More Comparison Using ELITE-SAG.} While existing works generally produce reasonable results, they often experience content ignorance or insufficient subject fidelity. This observation is especially obvious in complicated prompts, where the model has to comprehend complex relation between subjects. In contrast, ELITE-SAG achieves a balance between prompt consistency and subject fidelity.}
  \label{fig:supp_more_comparison_cat_1}
\end{figure*}

\begin{figure*}[!t]
  \includegraphics[width=0.9\textwidth]{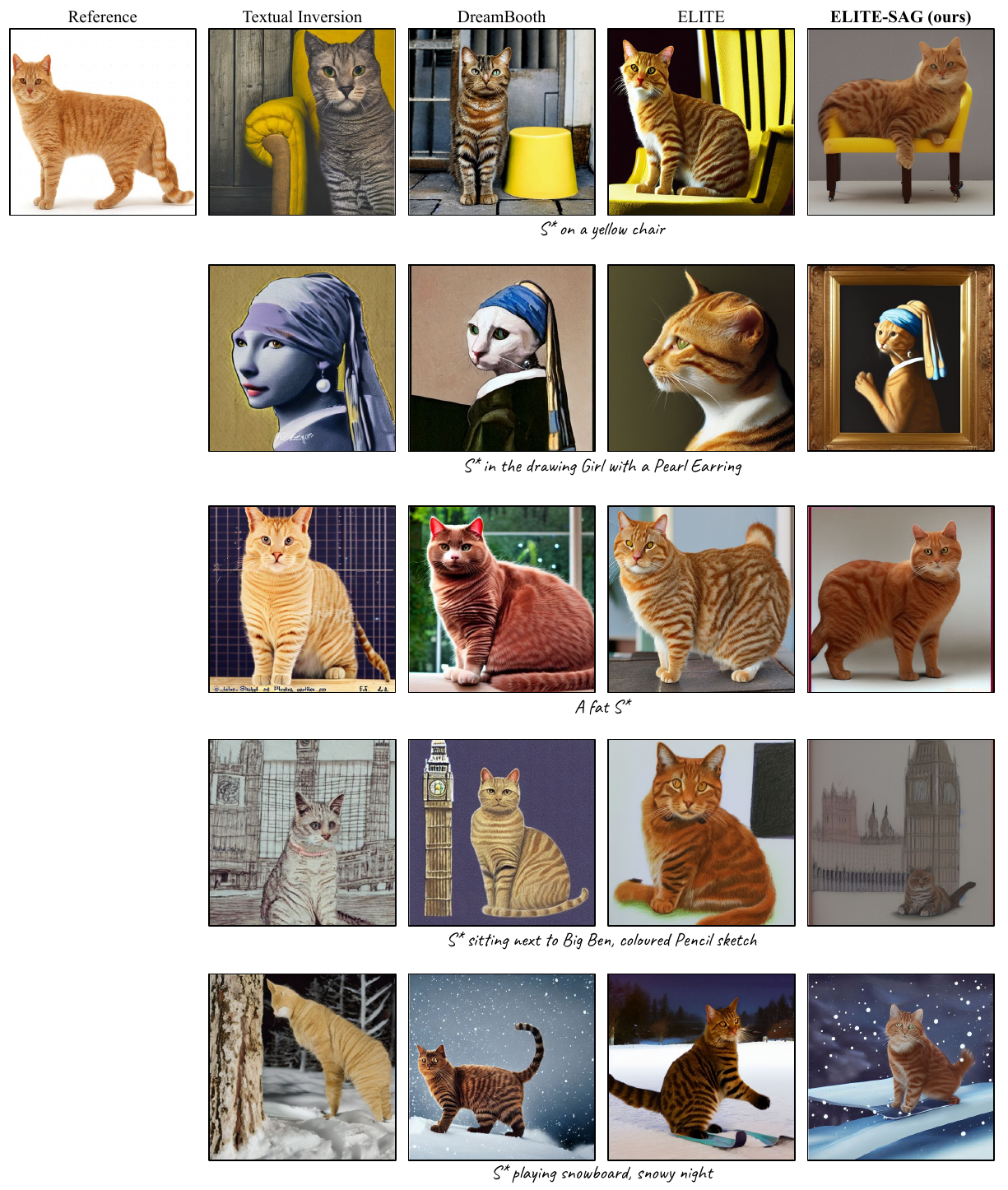}
  \caption{\textbf{More Comparison Using ELITE-SAG.} While existing works generally produce reasonable results, they often experience content ignorance or insufficient subject fidelity. This observation is especially obvious in complicated prompts, where the model has to comprehend complex relation between subjects. In contrast, ELITE-SAG achieves a balance between prompt consistency and subject fidelity.}
  \label{fig:supp_more_comparison_cat_2}
\end{figure*}

\section{Additional Results for Textual Inversion}
Additional results are shown in Fig.~\ref{fig:supp_ti}. With our SAG, Textual Inversion is able to produce results that are better align with the text descriptions.

\section{Additional Results for SuTI}
We provide additional results for applying SAG on SuTI in Fig.~\ref{fig:supp_suti}, without SAG, while identity is successfully preserved, the outputs often fail to capture the details specified by the text prompts. In contrast, text alignment is much better with SAG, without sacrificing identity. 

\section{Additional Results for DreamSuTI}
As depicted in Fig.~\ref{fig:supp_dreamsuti}, given a SuTI fine-tuned on a given style, our SAG leads to better style alignment while being able to preserve subject identity.

\section{Additional Comparison Using ELITE-SAG}
We provide additional comparison with DreamBooth \cite{ruiz2023dream}, Textual Inversion \cite{gal2023image}, and ELITE \cite{wei2023elite}. As shown in Fig.~\ref{fig:supp_more_comparison_dog_1} to Fig.~\ref{fig:supp_more_comparison_rabbit}, while existing works generally produce reasonable results, they often experience content ignorance or insufficient subject fidelity. This observation is especially obvious in complicated prompts, where the model has to comprehend complex relation between subjects. In contrast, ELITE-SAG achieves a balance between prompt consistency and subject fidelity. 

\end{document}